\documentclass[11pt]{article}

\usepackage[final]{acl}

\usepackage{times}
\usepackage{latexsym}
\usepackage{multirow}
\usepackage{booktabs} 
\usepackage{tabularx} 
\usepackage{kotex}
\usepackage{algorithm}
\usepackage{algpseudocode}

\usepackage[T1]{fontenc}

\usepackage[utf8]{inputenc}

\usepackage{microtype}

\usepackage{inconsolata}

\usepackage{graphicx}
\usepackage{xcolor}

\title{PsyProbe: Proactive and Interpretable Dialogue through User State Modeling for Exploratory Counseling}

\author{
    \textbf{Sohhyung Park\textsuperscript{1}}, 
    \textbf{Hyunji Kang\textsuperscript{2}}, 
    \textbf{Sungzoon Cho\textsuperscript{1$\dagger$}},
    \textbf{Dongil Kim\textsuperscript{2}}\\
    \textsuperscript{1} Department of Industrial Engineering, Seoul National University \\
    \textsuperscript{2} Department of Education, Seoul National University \\
    \normalsize\texttt{ sohhyung@bdai.snu.ac.kr}\\
    \normalsize\texttt{\{henge0821, zoon, dikimedu\}@snu.ac.kr}
    }

\begin{document}
\maketitle

\renewcommand{\thefootnote}{$\dagger$}
\footnotetext{Corresponding author.}
\renewcommand{\thefootnote}{\arabic{footnote}}

\begin{abstract}
Recent advances in large language models have enabled mental health dialogue systems, yet existing approaches remain predominantly reactive, lacking systematic user state modeling for proactive therapeutic exploration. We introduce PsyProbe, a dialogue system designed for the exploration phase of counseling that systematically tracks user psychological states through the PPPPPI framework (Presenting, Predisposing, Precipitating, Perpetuating, Protective, Impact) augmented with cognitive error detection. PsyProbe combines State Builder for extracting structured psychological profiles, Memory Construction for tracking information gaps, Strategy Planner for Motivational Interviewing behavioral codes, and Response Generator with Question Ideation and Critic/Revision modules to generate contextually appropriate, proactive questions. We evaluate PsyProbe with 27 participants in real-world Korean counseling scenarios, including automatic evaluation across ablation modes, user evaluation, and expert evaluation by a certified counselor. The full PsyProbe model consistently outperforms baseline and ablation modes in automatic evaluation. User evaluation demonstrates significantly increased engagement intention and improved naturalness compared to baseline. Expert evaluation shows that PsyProbe substantially improves core issue understanding and achieves question rates comparable to professional counselors, validating the effectiveness of systematic state modeling and proactive questioning for therapeutic exploration.
\end{abstract}

\section{Introduction}

Recent advances in large language models (LLMs) have led to growing interest in LLM-based counseling systems \cite{lawrence2024}. Counseling often involves deeply personal concerns and requires individualized attention as well as effective strategies for addressing emotional difficulties \cite{miller2013motivational,5pframework, hill2020helping}. Accordingly, deploying LLMs in therapeutic settings calls for careful consideration that goes beyond NLP techniques, incorporating therapeutic principles and perspectives.

One representative counseling model \cite{hill2020helping} identifies three stages of counseling: exploration, insight, and action. In this work, we focus on the exploration phase, which aims to deepen understanding by eliciting core issues underlying the user’s presenting problems \cite{hill2020helping}. To achieve this, counselors typically guide users to elaborate on their experiences beyond initial descriptions. Open-ended questioning plays a crucial role in facilitating this process \cite{Carona2021, williams2023use, Muntigl2023}. However, LLMs are fundamentally reactive, which can limit their ability to support proactive exploration.

Prior work has incorporated counseling theories into NLP systems, often leveraging Motivational Interviewing (MI) behavioral codes \cite{miller2013motivational} for automated classification \cite{rosas2017,tavabi2021analysis,gibson2022} and strategy-guided dialogue generation \cite{welivita2023boosting}. While these approaches provide structured guidance, they may not fully support proactive engagement during counseling interactions. Recent work such as \citet{hu-etal-2025-psyadvisor} utilizes questioning modules to guide exploration, but does not model user psychological states in sufficient detail for the counseling context, where diverse personal information continuously emerges throughout the dialogue. Additionally, research on Korean-language counseling systems remains limited. While the \citet{kim2025kmi} has adapted MI to Korean contexts, it relies on simulator-based synthetic data, limiting validation in real Korean counseling scenarios.

To address these limitations, we introduce \textbf{PsyProbe}, a proactive dialogue system that supports therapeutic exploration by explicitly modeling users’ psychological states through structured clinical formulation. PsyProbe integrates cognitive error signals with the PPPPPI framework (Presenting, Predisposing, Precipitating, Perpetuating, Protective, Impact) to represent and update core issues throughout the conversation. This formulation is operationalized through a State Builder that constructs the user state and a memory module that maintains and updates it over time. Using this evolving user state representation, the system plans MI-guided counseling strategies, identifies information gaps to address, and generates proactive follow-up questions that remain grounded in the counseling context. Our contributions are:

\begin{itemize}
\item We propose a structured user state modeling approach for exploratory counseling based on the PPPPPI framework, augmented with cognitive error signals to track evolving psychological core issues.
\item We develop a proactive question generation mechanism that combines gap-based Question Ideation with iterative Critic-and-Revision, supporting contextually appropriate and non-redundant exploration.
\item We conduct human evaluation with 27 participants and expert evaluation by a certified counselor, demonstrating that PsyProbe achieves question rates comparable to professional counselors while significantly improving user engagement and core issue understanding in real Korean counseling scenarios.
\end{itemize}

The full PsyProbe model consistently outperforms baselines across automatic evaluation settings. User and counselor evaluations show that PsyProbe attains question rates comparable to those of human counselors, while improving core issue understanding and the quality of probing questions. Compared to baseline systems, users also report higher engagement intention, suggesting the potential of systematic user state modeling and proactive questioning for therapeutic exploration.

\section{Related Work}

Advances in large language models \cite{GPT3, achiam2023gpt} have enabled mental health dialogue systems with improved contextual understanding and domain adaptation \cite{yuan2025}. Early systems like Woebot \cite{fitzpatrick2017} implemented cognitive behavioral therapy through rule-based approaches with predefined dialogue flows \cite{Denecke2021}. Subsequent work leveraged commonsense knowledge to enhance emotional responsiveness \cite{deng2023}, expanding applications to clinical data analysis \cite{taylor2024model} and automated coaching \cite{ong2024advancing}. For Korean contexts, initial efforts focused on psychiatric counseling \cite{oh2017} and role-specific system design \cite{bae2022building}, as well as empathetic chatbots \cite{palbot}.  Growing interest in therapeutic dialogue has spurred research on incorporating Motivational Interviewing (MI) techniques \cite{miller2013motivational} into conversational systems, with studies focusing on automated classification of MI behavioral codes \cite{rosas2017,gibson2022,tavabi2021analysis} and strategy-guided approach \cite{welivita2023boosting}. In Korean contexts, \citet{kim2025kmi} constructed an MI-annotated Korean counseling dataset.

Despite these advances, existing systems remain predominantly reactive, lacking systematic tracking of user psychological states or strategic planning for therapeutic exploration. In counseling practice, the three-stage framework of exploration, insight, and action \cite{hill2020helping} emphasizes the importance of proactive techniques throughout the therapeutic process. Open-ended questioning proves especially critical for structuring thoughts and facilitating deeper exploration \cite{hill2020helping, Carona2021}, with empirical evidence supporting its effectiveness in therapeutic engagement \cite{williams2023use, Muntigl2023}. Recently, \citet{hu-etal-2025-psyadvisor} introduced a modular approach incorporating retrieval of counseling theories and cases to inform strategic questioning.

\begin{figure*}[t]
    \centering
    \includegraphics[width=0.95\linewidth]{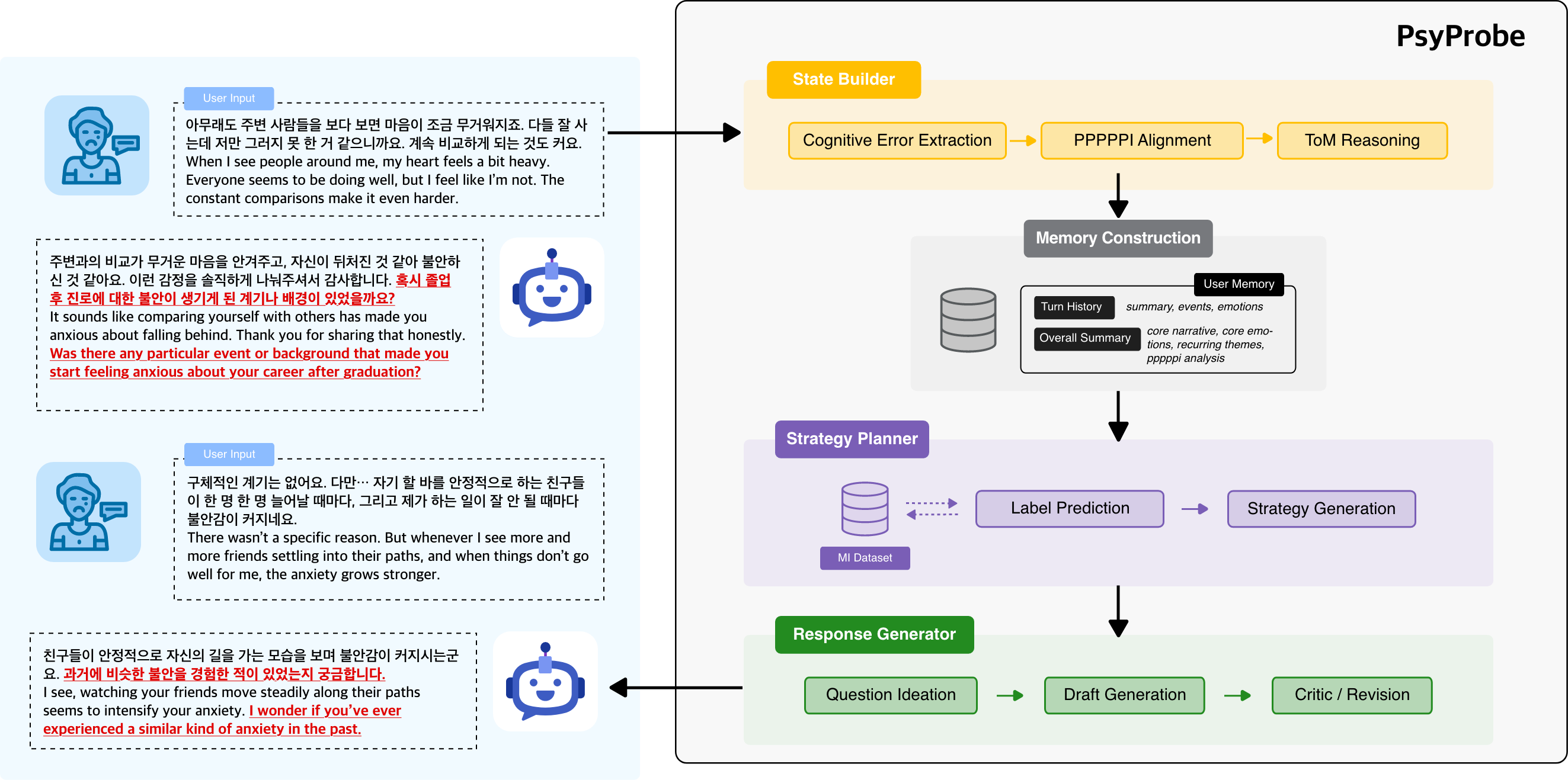}
    \caption{Overview of the PsyProbe system architecture.}
    \label{fig:overall_framework}
\end{figure*}

\section{PsyProbe}

PsyProbe is a proactive dialogue system for the exploration phase of counseling, an early-stage process where presenting concerns are gradually elaborated into a clearer understanding of the client’s core issues. We define core issues through the PPPPPI framework (Presenting, Predisposing, Precipitating, Perpetuating, Protective, Impact) augmented by cognitive error signals. 

PsyProbe consists of four modules: State Building, Memory Construction, Strategy Planning, and Response Generation.  The State Builder extracts these structured representations from user utterances, Memory Construction tracks evolving gaps in understanding, and the Strategy Planner predicts MI behavioral codes to guide response generation. The Response Generator employs Question Ideation to generate targeted questions based on PPPPPI gap scores, and Critic and Revision to iteratively refine drafts, ensuring contextually appropriate, proactive questioning that deepens exploration.

\subsection{State Builder}
To extract core issues from presenting problems, we must first define what a "core issue" is. The State Builder addresses this through three stages: (1) \textbf{Cognitive Error Extraction} identifies distorted thinking patterns, (2) \textbf{PPPPPI Alignment} maps content to clinical formulation slots (Presenting, Predisposing, Precipitating, Perpetuating, Protective, Impact), and (3) \textbf{Theory-of-Mind (ToM) Reasoning} infers underlying beliefs, desires, and intentions. These stages transform unstructured conversational input into structured psychological profiles that guide subsequet modules. Further implementation details are reported in Appendix~\ref{app:state_builder_details}.

\subsubsection{Cognitive Error Extraction}
\label{sec:cog_error_extraction}
Cognitive errors are maladaptive thinking patterns that contribute to emotional disorders such as depression and anxiety, correlate with symptom severity, and serve as key therapeutic targets \citep{beck1976cognitive,beck1979cognitive}. 
\citet{beck1976cognitive} identified several primary types of cognitive distortions that characterize depressive cognition, and \citet{lefebvre1981cognitive} empirically reconceptualized these into four broader categories—catastrophizing, overgeneralization, personalization, and selective abstraction. In this stage, the model identifies cognitive errors from user input based on these categories. 

In particular, selective abstraction was defined through explicit “should” and “must” expressions. These expressions, which can be categorized as instances of selective abstraction, have been shown to strongly reflect depressive and self-critical thinking patterns, being more frequent in depressive cognition and significantly associated with depression severity \citep{ellis1980rational,blake2016}. Although sub-patterns such as jumping to conclusions or emotional reasoning can also be clustered under selective abstraction \citep{blake2016}, they often require contextual inference across multiple utterances to be reliably detected. Therefore, we focus on should/must expressions that are both prominent and explicitly observable within short user utterances.

The types of cognitive errors and their representative expressions used in this stage can be found in Appendix~\ref{app:cognitive_error_definition}. The model is prompted to detect these cognitive errors and extract the corresponding evidence spans within each user utterance, which are then used as signals in the next stage to define the user's core problem.

\subsubsection{PPPPPI Alignment}

The Five Ps framework offers a structured lens for understanding psychological difficulties by organizing complex personal, social, and contextual information beyond diagnostic categories. Within this formulation, the Presenting problem corresponds to the user’s expressed difficulties or symptoms in their utterances, which are contextualized through Predisposing, Precipitating, Perpetuating, and Protective factors that also encompass personal strengths buffering distress \citep{5pframework}. Although this framework structure causal and protective dimensions of distress, clinical and diagnostic standards further emphasize functional impairment as a core indicator, reflecting the extent to which symptoms disrupt everyday life \cite{american2000diagnostic, ustun2010measuring, kroenke2001phq9}. Therefore, we incorporate an Impact dimension into the core problem formulation, capturing how users’ problems affect daily functioning across multiple domains.

In this stage, the system performs evidence-grounded PPPPPI slot filling using the cognitive error flags produced in the previous stage and the current user utterance. The definitions of the PPPPPI slots used in this stage can be found in Appendix~\ref{app:pppppi_alignment_definition}. For each utterance, the model detects any PPPPPI elements that are explicitly supported by the text and extracts the corresponding evidence spans. If no textual evidence exists, the slot is left empty. 

\subsubsection{ToM Reasoning}

Theory of Mind (ToM) refers to the capacity to infer another person’s mental states, including beliefs, desires, and intentions \citep{Premack_Woodruff_1978}. In our setting, we use the concept of ToM to infer how the user interprets events (beliefs), what the user wants (desires), and the likely interactional intent behind the utterance (intent) so that responses are targeted and aligned with the user’s needs.

In this stage, the module takes the PPPPPI outputs, including evidence spans, as input. It first derives beliefs and desires that are directly supported by text, written as short phrases rather than copying the original sentences. The module then selects a single intent label from the four processes of Motivational Interviewing (MI): Engaging, Focusing, Evoking, and Planning \citep{miller2013motivational}. We use “intent” to denote the utterance’s immediate interactional need, such as empathic support, clarification of the topic or goal, exploration of change, or readiness to agree on next steps. The selected label is chosen to align with the inferred beliefs and desires and is supported by the same evidence. The Strategy Planner then uses this label to map to MI micro-skills. Chain of thought prompting is used internally to reason step by step \citep{cot}. This stepwise beliefs–desires–intent procedure produces a concise cue from PPPPPI-grounded textual evidence that indicates the user’s likely intent and serves as a hint for the counselor’s next move.

\subsection{Memory Construction}

The system memory has two layers: (1) \textit{turn history}, a lightweight record of recent dialogue that stores keywords, events with context, and emotions with triggers; and (2) \textit{overall summary}, a clinical snapshot consisting of a PPPPPI analysis and a summary (core narrative, core emotion, recurring themes). The overall summary serves as the system’s current formulation of the core problem. The full input context, schema and examples are provided in Appendix~\ref{app:memory_construction}.

\paragraph{Turn history construction}
The system builds the turn history from the user's most recent utterances and recent dialogue history, extracting and storing keywords, event–context pairs, and emotion–trigger pairs. Entries are written as short phrases. 

\paragraph{PPPPPI analysis update}

The system updates the six PPPPPI entries based on the current turn history, the State Builder’s latest PPPPPI outputs, and the existing PPPPPI analysis. Each entry is stored as concise text and updated based on a conservative, evidence-grounded policy. When direct textual evidence is absent, prior content is left unchanged, and any reasonable inference is explicitly marked. 

\paragraph{Summary update}
The system integrates the turn history, the updated PPPPPI analysis, and the existing summary, and updates core narrative, core emotion, and recurring themes as concise text.

\subsection{Strategy Planner}
The Strategy Planner guides the system’s next turn through two stages: (1) \textbf{Label Prediction} predicts motivational interviewing (MI) behavioral codes as actionable guidance for the next system turn, and (2) \textbf{Strategy Generation} converts the predicted codes into a concrete dialogue strategy that specifies the therapeutic goal, key content points to address, and stylistic constraints for each selected speech act. Detailed implementation and prompting templates are provided in Appendix~\ref{app:strategy_planner_details}.

\subsubsection{Label Prediction}
Major MI skills such as OARS are frequently employed in the exploratory phase, providing actionable next-turn options \citep{miller2013motivational, hill2020helping}. We predict two MI behavioral codes using few-shot prompting with Korean examples from the KMI corpus \citep{kim2025kmi}, which provides counseling sessions annotated with professional therapists' behavioral choices.

The prediction runs in two rounds. In round one, it returns one code from this list: Simple Reflection, Complex Reflection, Open Question, Closed Question, Affirm, Give Information, Advise, General, together with a brief Korean rationale explaining why the next agent turn should use that code. In round two, it removes the first-picked label and, for Questions and Reflections, removes both subtypes to prevent selecting another from the same category.

\subsubsection{Strategy Generation}
Following label prediction, the system constructs a dialogue strategy that specifies how each selected speech act should be realized in the response. 

The model utilizes the predicted labels, ToM signals, and memory information to generate detailed act plans for each selected speech act. Each act plan specifies the therapeutic goal, key content points to address, and stylistic hints for delivery. For instance, an Open Question act plan might define its goal as exploring functional impact, include key points such as "exploring daily life contexts" and "clarifying functional impact," with style hints like "maintain empathic tone" and "avoid judgmental interpretations."

This structured planning ensures that the final response is purposeful, contextually grounded, and therapeutically sound. By explicitly defining the goal and execution strategy for each speech act before response generation, the system bridges high-level behavioral codes with concrete, empathic dialogue that aligns with therapeutic principles and the user's current psychological state.

\subsection{Response Generator}
The Response Generator produces the final counselor turn through three stages: (1) \textbf{Question Ideation} generates candidate probing questions using gap scores over PPPPPI slots as signals of what remains under-specified in the current formulation,
 (2) \textbf{Draft Generation} composes an initial response by combining the planned speech acts with dialogue-grounded content while maintaining an empathic and concise counseling tone, and (3) \textbf{Critic and Revision} verifies contextual appropriateness and redundancy and revises the question component when needed to produce a therapeutically coherent and exploration-supportive response. Additional implementation details are provided in Appendix~\ref{app:response_generator_details}.

\subsubsection{Question Ideation}
The Question Ideation stage generates candidate questions that target information gaps in the PPPPPI framework. The system computes a gap score for each slot using a heuristic that assigns weights based on whether the slot content is missing, weak in evidential support, or outdated. Slots with higher gap scores are prioritized for exploration. For each high-priority slot, the module generates candidate questions designed to gather specific types of information. Additional details on the gap scoring heuristic are provided in Appendix~\ref{app:ppppi_gap_scoring}.

However, not all slots are treated equally. The exploration phase of counseling aims to deeply understand the client's presenting problems rather than prematurely shift toward solutions. Since questions about Protective factors tend to focus on problem-solving and coping strategies, they risk redirecting the conversation away from exploration before the core issues are fully clarified. Therefore, the system generates Protective questions conservatively, only when the Protective gap is substantial or when the user's recent utterance contains relevant cues indicating readiness to discuss strengths or resources. This constraint prevents the system from imposing external problem-solving frameworks before the user's difficulties are thoroughly explored. 

The module receives inputs including current PPPPPI analysis, recent dialogue turns, keywords from turn history, and top-k slots with highest gap scores. These inputs enable the system to balance both long-term thematic understanding and immediate conversational relevance when generating questions. The output is a ranked list of candidate questions, each tagged with its target slot, intent, and gap score. Detailed input specifications are provided in Appendix~\ref{app:question-ideation}.

\subsubsection{Draft Generation}
Draft Generation produces an initial response by integrating the selected speech acts with appropriate content from the dialogue context. 

The system applies different generation strategies depending on the selected speech act label, enforcing additional rules for Reflections and Open Questions. Across all speech acts, the prompt includes common constraints such as maintaining an empathic, non-judgmental tone and keeping responses concise, typically two to three sentences with a maximum of four. The draft combines the primary and secondary speech acts into a coherent turn that feels natural rather than formulaic.

The output is a draft response that aims to balance therapeutic appropriateness with conversational fluidity, ready for evaluation by the Critic module.

\subsubsection{Critic and Revision}
The Critic module evaluates the draft response for contextual appropriateness, redundancy, and therapeutic soundness. 

The evaluation focuses on three key dimensions. First, the critic assesses the presence and necessity of a question. If the draft does not contain a question, the critic determines whether adding one would benefit the exploratory process. If the draft contains a question, the critic evaluates whether it is appropriate for the current conversational state and supports continued exploration. Second, the critic checks whether the draft aligns with the user's most recent utterance and the broader conversational flow, identifying any mismatches in tone, content, or focus. Third, it detects redundancy by comparing the draft against recent agent turns, particularly checking whether the draft question or reflection pattern repeats content from recent responses.

Based on this evaluation, the critic generates a set of revision operations for the question component, which can be \textit{keep}, \textit{add}, \textit{replace}, or \textit{remove}, accompanied by a rationale explaining the reasoning behind each operation. If no question is present but exploration would benefit from one, the critic may specify add along with guidance on which gap to prioritize.

The Refiner then applies these operations to produce the final response. If the operation is keep, the draft is returned as is. If replace or add is specified, the system selects an appropriate question from the candidate pool, ensuring alignment with the revised strategy. If remove is specified, the question is deleted and the response relies solely on the reflection or other speech acts. The output is a final agent response that has been verified for contextual fit, novelty, and therapeutic appropriateness before delivery to the user. 

\section{Experiment}
\subsection{Experimental Design}
\subsubsection{Procedure}
The experiment involved a short demographic survey, a pre-session questionnaire, a counseling session, and a post-session evaluation. All interactions were conducted through a web-based interface. We recruited 27 participants who each completed three counseling sessions in random order, corresponding to three modes: GPT, PsyProbe (Ours), and a professionally certified human counselor. The study received ethical approval from the Institutional Review Board (IRB) of Seoul National University (IRB No.~2509/004-009).

\paragraph{Demographic Survey} 
Participants first completed a brief demographic form collecting basic information such as age, gender, current occupation, and their frequency and purpose of LLM use. We recruited 27 participants, and their demographics are reported in Appendix~\ref{app:demographics}.

\paragraph{Pre-session questionnaire}
Each participant was asked to describe three personal concerns and the emotions associated with each before the counseling sessions. These inputs were later used as the initial prompts for three separate counseling sessions. To ensure consistency, all participants were asked to maintain the same discussion topic throughout the three sessions.

\paragraph{Counseling session}
Each participant engaged in three counseling sessions corresponding to the GPT, \textit{PsyProbe}, and human counselor modes. 
In all sessions, the participant discussed the same personal topic described in the pre-session questionnaire. Both model-based counselors simulated the role of a human counselor by providing empathetic responses based on the participant’s input. The GPT-based counselor was implemented as a basic model prompted with Rogers' therapeutic conditions of client-centered counseling, which emphasize empathy, congruence, and unconditional positive regard \citep{rogers1957necessary, rogers1959theory}. Each session lasted up to 20 minutes or ended earlier if the participant indicated completion.

\paragraph{Post-session evaluation}
After the session, participants completed a short questionnaire assessing their emotional changes compared to the pre-session state and their perception of the chatbot’s performance across multiple items.

\subsubsection{Implementation Details}
We implemented PsyProbe using two large language models: GPT-4o and Claude-3.5-Haiku. Both implementations use identical system architecture and prompting strategies across all modules. The Strategy Planner uses 3 few-shot examples from the KMI Dataset \citep{kim2025kmi} for MI behavioral code prediction. For automatic evaluation , we report results for both model implementations. For human evaluation, we used the GPT-4o implementation.
\subsection{Evaluation Metric}

\subsubsection{Automatic Evaluation}
Automatic evaluation compares model-generated responses with human counselor responses collected from the human-counselor sessions in our study. For each user utterance, we generate a model response and treat the corresponding human counselor reply as the reference, measuring content similarity using ROUGE (R-1, R-2, R-L), BLEU (B-1, B-2, B-3, B-4), and BERTScore-F1.
For \textit{PsyProbe}, we report results for five configurations: the baseline model without additional modules, the model without the State Builder, without the Strategy Planner, without the Question Ideation and Critic components, and the full model including all modules.

\subsubsection{User Evaluation}
Following evaluation dimensions frequently adopted in human assessments of conversational agents \citep{Adiwardana2020,bae2022building,chen-etal-2023-places,palbot}, we asked participants to rate the chatbot’s performance on five aspects: \textit{Coherence}, \textit{Fluency}, \textit{Naturalness}, \textit{Empathy}, and \textit{Safety}. All items were evaluated on a 5-point Likert scale (1--5). 

Participants also rated \textit{Engagement Intention}, which reflects whether the conversation encourages users to continue the dialogue and further explore their concerns, on a 3-point scale (0 = not willing, 1 = neutral, 2 = willing).

\begin{table*}[!t]
\centering
\caption{Automatic evaluation results for \textbf{PsyProbe} across different ablation settings, measured by ROUGE (R-1/2/L), BLEU (B-1/2/3/4), and BERT-F1; bold indicates the best score within each model.}

\resizebox{0.9\textwidth}{!}{%
\begin{tabular}{l l c c c c c c c c}
\toprule
Model & Mode & R-1 & R-2 & R-L & B-1 & B-2 & B-3 & B-4 & BERT-F1 \\
\midrule
\multirow{5}{*}{GPT-based} 
   & Baseline & 0.2127 & 0.0636 & 0.1372 & 0.1754 & 0.0928 & 0.0567 & 0.0377 & 0.5428 \\
   & w/o SB & 0.2181 & 0.0690 & 0.1492 & 0.1900 & 0.1039 & 0.0654 & 0.0446 & 0.5477 \\
   & w/o SP & \textbf{0.2280} & 0.0724 & 0.1516 & 0.1956 & 0.1068 & 0.0661 & 0.0447 & 0.5525 \\
   & w/o QI/C & 0.2223 & 0.0695 & 0.1473 & 0.1931 & 0.1050 & 0.0662 & 0.0451 & 0.5470 \\
   & Full & \underline{0.2277} & \textbf{0.0729} & \textbf{0.1543} & \textbf{0.1968} & \textbf{0.1082} & \textbf{0.0670} & \textbf{0.0452} & \textbf{0.5540} \\
\midrule
\multirow{5}{*}{Claude-based} 
   & Baseline & 0.1776 & 0.0408 & 0.1053 & 0.1276 & 0.0593 & 0.0326 & 0.0205 & 0.5242 \\
   & w/o SB & 0.2037 & 0.0513 & 0.1248 & 0.1693 & 0.0821 & 0.0457 & 0.0286 &  0.5270 \\
   & w/o SP & 0.2051 & 0.0530 & 0.1271 & 0.1688 & 0.0835 & 0.0472 & 0.0297 & 0.5358 \\
   & w/o QI/C & 0.1948 & 0.0444 & 0.1186 & 0.1622 & 0.0749 & 0.0417 & 0.0265 & 0.5211  \\
   & \textbf{Full} & \textbf{0.2096} & \textbf{0.0550} & \textbf{0.1322} & \textbf{0.1731} & \textbf{0.0858} & \textbf{0.0494} & \textbf{0.0321} & \textbf{0.5378} \\
\bottomrule
\end{tabular}%
}

\label{tab:automatic}
\end{table*}

\subsubsection{Expert Evaluation}
To examine how well the model supports the early exploration phase of counseling, we conducted an expert evaluation of the model-generated outputs.

Since these aspects are highly dependent on the ongoing interaction and session context, the certified counselor who conducted each session rated the model-generated outputs on two dimensions, \textit{Core Issue Understanding} (CIU) and \textit{Probing Question Quality} (PQQ), along with the \textit{Question Rate}.

\textit{Core Issue Understanding} (CIU) was rated on a 5-point Likert scale (1--5), assessing how well the model identified the client’s underlying core issue. 

\textit{Probing Question Quality} (PQQ) was rated on a 3-point scale (0--2), assessing the relevance and depth of the model’s probing questions, where 0 indicates unnatural or irrelevant questions, 1 indicates acceptable but unhelpful questions, and 2 indicates natural and appropriate questions.

\section{Results}
\subsection{Automatic Evaluation}

Table ~\ref{tab:automatic} presents automatic evaluation results comparing model-generated responses with human counselor responses, measured by lexical overlap (ROUGE, BLEU) and embedding-based similarity (BERT-F1). We evaluate five modes: baseline, ablations removing State Builder (w/o SB), Strategy Planner (w/o SP), Question Ideation and Critic (w/o QI/C), and the full system.
The full PsyProbe system demonstrates improvements over baseline across both implementations. For GPT-based models, lexical overlap metrics show gains with a small increase in embedding-based similarity. Claude-based models exhibit larger improvements, suggesting that structured modules provide greater benefit when baseline performance is lower.

Ablation results reveal that State Builder and Question Ideation/Critic modules contribute most substantially across both implementations. In GPT-based models, removing State Builder yields the lowest performance among ablations, while removing QI/C also degrades performance. Claude-based models show a similar pattern, with QI/C removal causing the largest drop and State Builder removal also degrading results across metrics. These results suggest that both modules play a central role in supporting the exploration phase by grounding the conversation in an explicit user state and generating probing questions that encourage users to further elaborate on their concerns. Strategy Planner shows relatively lower contribution, with w/o SP modes performing similarly to Full models in both implementations, which may reflect the limited sensitivity of n-gram overlap metrics to paraphrasing differences (Appendix~\ref{app:sp_examples}).

While the full model consistently achieves the highest BERT-F1 scores across both implementations, the improvement magnitude remains small. Since baseline models were prompted with Rogers' client-centered principles to ensure fair comparison in the counseling context, they already generate semantically appropriate responses. Additionally, when addressing the same general counseling topics and contexts, the core therapeutic meaning naturally remains similar across different response formulations. PsyProbe primarily enhances how responses are expressed rather than fundamentally altering their semantic direction.

However, these automatic metrics have inherent limitations. While they capture improvements in lexical alignment with human responses, the absolute gains are relatively small. Additionally, counseling responses can vary substantially across individuals and sessions, making reference-based metrics less reliable. More critically, these metrics cannot assess proactive questioning quality, therapeutic appropriateness, or the depth of exploration. These aspects are more directly evaluated through user and expert assessments presented in subsequent sections.

\subsection{User Evaluation}

Table~\ref{tab:human-eval} shows the user evaluation results comparing GPT, PsyProbe, and human counselors across six dimensions. PsyProbe demonstrates the most substantial improvement in Naturalness, approaching human counselor levels. This suggests that the combination of strategic questioning and reflections creates a more natural conversational flow than purely reactive responses. PsyProbe also shows modest gains in Coherence and Safety, maintaining high performance across these dimensions.  Fluency is slightly lower in PsyProbe compared to GPT, though both remain high. While both systems apply the same length constraints, GPT tends to generate longer reactive responses, whereas PsyProbe enforces additional structural requirements for explicit conversational acts and targeted question formulation, which may account for this minor gap. Empathy shows only marginal improvement, with both PsyProbe and GPT remaining substantially below human counselors. This gap suggests that simple reflection strategy instructions are insufficient to capture the diversity and depth of human empathic responses. Human counselors demonstrate more varied reflection patterns and can sustain conversational flow through reflection alone. PsyProbe's primary focus on proactive questioning may limit the development of rich, multifaceted reflection strategies.

Engagement Intention shows significant improvement, with PsyProbe scoring 1.19 compared to GPT's 0.33, approximately 3.6 times higher. This indicates that systematic modeling of user state through the PPPPPI framework and the resulting proactive questioning significantly increase users' willingness to continue the conversation. Participants' qualitative responses in the post-session evaluation further support this observation, with some describing baseline GPT responses as prematurely wrapping up the conversation and leaving fewer natural openings for follow-up.

\begin{table}[t]
\small
\caption{User evaluation results comparing GPT, \textbf{PsyProbe}, and human counselors. 
C: Coherence, F: Fluency, N: Naturalness, E: Empathy, S: Safety, EI: Engagement Intention.}
\centering
\small
\begin{tabular}{lcccccc}
\toprule
Mode & \textbf{C} & \textbf{F} & \textbf{N} & \textbf{E} & \textbf{S} & \textbf{EI} \\
\midrule
GPT & 4.10 & 4.19 & 2.87 & 3.77 & 4.45 & 0.33 \\
PsyProbe & 4.20 & 4.08 & 3.72 & 3.84 & 4.50 & 1.19 \\
Human & 4.58 &  4.39 & 4.19 & 4.52 &4.55 & 1.59 \\
\bottomrule
\end{tabular}

\label{tab:human-eval}
\end{table}

\subsection{Expert Evaluation}

Table~\ref{tab:expert-eval} presents expert evaluation results that directly assess the system's capacity for proactive exploration. The most striking result appears in Question Rate (QR), where PsyProbe achieves 0.815 compared to GPT's 0.263, a rate comparable to human counselors at 0.830. This demonstrates that the combination of Question Ideation and Critic/Revision modules successfully increases questioning frequency to a level approaching professional therapeutic practice.
\begin{figure}[htpb]
\centering
\includegraphics[width=0.95\columnwidth]{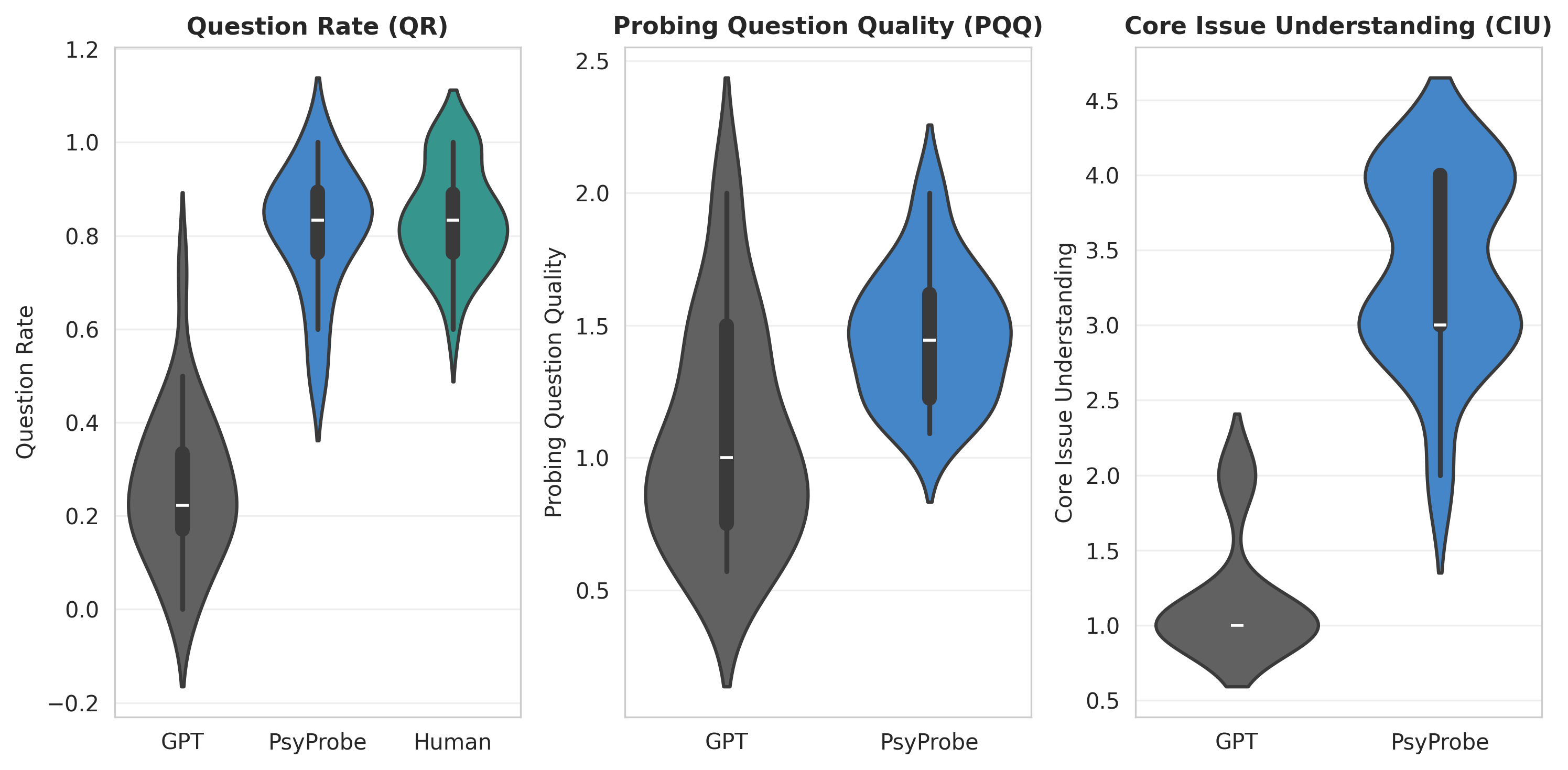}
\caption{Expert evaluation results across three metrics. }
\label{fig:expert-eval}
\end{figure}

Probing Question Quality (PQQ) similarly shows improvement, demonstrating that questions generated are not only more frequent but also of higher quality, falling above the "acceptable" threshold and approaching the "natural and appropriate" level. The refinement process appears effective in selecting questions that meaningfully advance exploration of PPPPPI gaps.

Core Issue Understanding (CIU) shows substantial improvement, with PsyProbe scoring 3.370 compared to GPT's 1.153 on a 5-point scale. This validates the effectiveness of the State Builder and Memory Construction modules in systematically tracking and synthesizing the user's presenting problems within the PPPPPI framework. The expert rater judged that PsyProbe maintained coherent understanding of core issues throughout the dialogue, whereas GPT struggled to identify and maintain focus on central therapeutic themes. However, the absolute score of 3.37 suggests room for improvement. The 20-minute time constraint on each session may contribute to limitations in exploration depth, and opportunities remain for enhancing core issue understanding in future iterations.

\begin{table}[!htpb]
\caption{Expert evaluation results comparing GPT, \textbf{PsyProbe}, and human responses. 
QR: Question Rate, proportion of question-type utterances, CIU: Core Issue Understanding, PQQ: Probing Question Quality.}
\small
\centering
\begin{tabular}{lccc}
\toprule
Mode & \textbf{QR} & \textbf{CIU} & \textbf{PQQ} \\
\midrule
GPT & 0.263 & 1.153 & 0.951 \\
PsyProbe (Ours) & \underline{0.815} & \textbf{3.370} & \textbf{1.440} \\
Human & \textbf{0.830} & -- & -- \\
\bottomrule
\end{tabular}

\label{tab:expert-eval}
\end{table}


\section{Conclusion}
We introduce \textbf{PsyProbe}, a proactive dialogue system for exploratory counseling that systematically models user psychological states through the PPPPPI framework. PsyProbe integrates a \textit{State Builder} for cognitive error detection and structured clinical formulation, \textit{Memory Construction} for tracking information gaps, and a \textit{Response Generator} that produces proactive questions through \textit{Question Ideation} and iterative \textit{Critic/Revision}. Through this modular design, PsyProbe generates contextually appropriate, proactive questions that deepen therapeutic exploration. Automatic evaluation shows that PsyProbe consistently outperforms GPT-based baselines, with ablation results indicating that the State Builder and the Question Ideation with Critic/Revision modules contribute most to these gains. In expert evaluation, PsyProbe demonstrates substantial improvements in proactive questioning quality and core issue understanding, while achieving question rates comparable to professional counselors. Participants reported significantly higher engagement intention than GPT-based baselines, with higher ratings for naturalness, empathy, safety, and coherence. While the system shows promising results in naturalness and questioning strategy, empathy remains an area for improvement, suggesting that richer reflection strategies beyond simple instructions are needed. Future work should focus on enhancing empathic responsiveness and extending the system to support insight and action phases of counseling, creating a comprehensive therapeutic dialogue framework.

\section*{Limitations}

This work has several limitations that should be considered. First, while we demonstrate practical improvements without fine-tuning, the system relies heavily on multiple LLM calls across various modules, which may limit scalability and increase latency in real-world deployment. The modular architecture, though interpretable, requires careful prompt engineering and may be sensitive to variations in LLM performance. Second, the lack of established datasets and structured frameworks for evaluating proactive counseling systems poses challenges for systematic development and comparison. Our PPPPPI-based formulation and cognitive error detection provide an insightful approach to structuring therapeutic dialogue, but more research is needed to develop standardized evaluation protocols and annotated corpora that capture the nuances of exploratory counseling. Third, the 20-minute session constraint and limited participant pool (27 participants) may not fully capture the diversity of counseling scenarios and client needs. Finally, while our system shows promise in the exploration phase, extending it to support insight and action phases requires further investigation into how psychological state tracking and strategic questioning should adapt across different therapeutic stages.

\section*{Acknowledgments}

This work was supported by the Ministry of Education of the Republic of Korea and the National Research Foundation of Korea (NRF-2020S1A3A2A02103411), and by the BK21 FOUR Program (Education and Research Center for Industrial Innovation Analytics) funded by the Ministry of Education, Korea. During manuscript preparation, we used Claude and GPT for language assistance, including English writing support and Korean--English translation. All AI-generated content was carefully reviewed for accuracy and appropriateness. 

\bibliography{custom}

\newpage
\appendix

\section{State Builder Schema}
\label{app:state-builder}

This section summarizes the definitions used in the State Builder, including cognitive error definitions and the PPPPPI slot definitions for evidence-grounded formulation.

\subsection{Cognitive Error Extraction}
\label{app:cognitive_error_definition}

Table~\ref{tab:cognitive_errors} summarizes the four cognitive error categories and their representative expressions used in the Cognitive Error Extraction stage of the State Builde. These categories are based on \citeposs{lefebvre1981cognitive} empirical reconceptualization of \citeposs{beck1976cognitive} original cognitive distortions. Each category represents a distinct pattern of maladaptive thinking that contributes to emotional disorders such as depression and anxiety.

\begin{table*}[htpb]
\centering
\small
\caption{Cognitive error categories and representative expressions; the four categories follow \citeposs{lefebvre1981cognitive} reconceptualization of \citeposs{beck1976cognitive} original distortions.}
\begin{tabularx}{\textwidth}{X X X}
\toprule
\textbf{Category} & \textbf{Definition} & \textbf{Example expressions} \\
\midrule
\textbf{Catastrophizing} & Interpreting events as disastrous or the worst possible outcome. & ``disaster,'' ``ruined,'' ``it's over'' \\
\textbf{Overgeneralization} & Making broad or absolute conclusions from limited experiences. & ``always,'' ``never,'' ``everyone,'' ``no one'' \\
\textbf{Personalization} & Assigning excessive self-blame or responsibility for negative outcomes. & ``my fault,'' ``I ruined it,'' ``I'm the problem'' \\
\textbf{Selective abstraction (Should statements)} & Focusing on rigid moral or duty-based expectations. & ``must,'' ``should,'' ``have to,'' ``shouldn't'' \\
\bottomrule
\end{tabularx}
\label{tab:cognitive_errors}
\end{table*}

\subsection{PPPPPI Alignment}
\label{app:pppppi_alignment_definition}

Table~\ref{tab:pppppi-slots} defines the six PPPPPI dimensions used for clinical formulation. Each slot captures a distinct aspect of the user's psychological state and contextual factors.

\begin{table*}[htpb]
\centering
\small
\caption{Definitions of PPPPPI slots for clinical formulation.}
\begin{tabularx}{\textwidth}{lX}
\toprule
\textbf{Slot} & \textbf{Definition} \\
\midrule
\textbf{Presenting} & Current expressed problems, emotions, or complaints (e.g., feeling overwhelmed, anxious, depressed, distressed, self-blame) \\
\textbf{Precipitating} & Recent triggering factors, events, or temporal cues (e.g., yesterday/today/recently, conflicts, arguments, being criticized) \\
\textbf{Perpetuating} & Factors that maintain or worsen the problem, including repetitive patterns, cognitive distortions (e.g., always/constantly, catastrophizing, self-blame, overgeneralization, should statements) \\
\textbf{Predisposing} & Long-standing vulnerabilities or background factors (e.g., longstanding tendencies, personality traits, environmental background). \textit{Left empty if no evidence exists.} \\
\textbf{Protective} & Protective factors, strengths, or support systems (e.g., seeking/receiving help, expressing values/meaning/strengths, supportive people or resources) \\
\textbf{Impact} & Functional impact on daily life (e.g., sleep disturbance, impaired academic/work performance, interpersonal difficulties, avoidance, fatigue with \textit{specific effects}) \\
\bottomrule
\end{tabularx}

\label{tab:pppppi-slots}
\end{table*}

\section{PPPPI Gap Scoring for Question Ideation}
\label{app:ppppi_gap_scoring}

To guide proactive questioning during the exploration phase, PsyProbe maintains a structured representation of the user’s core issue using PPPPPI slots and estimates which parts remain under-specified.
Specifically, for each PPPPPI slot, we compute a \textit{gap score} that summarizes how incomplete the current slot state is.
As shown in Algorithm~\ref{alg:ppppi_gap_simple}, the gap score is computed as a weighted sum of four binary signals: (i) whether the slot content is missing, (ii) whether the evidence supporting the slot is weak (e.g., no explicit span or low confidence), (iii) whether the provenance is weak (e.g., lacking strong supporting signals), and (iv) whether the slot was not updated in the current turn.
In our experimental setting, we fixed the heuristic weights based on pilot observations and feedback from a certified counselor ($w_{\texttt{content}}{=}0.40$, $w_{\texttt{evidence}}{=}0.45$, $w_{\texttt{prov}}{=}0.20$, and $w_{\texttt{recency}}{=}0.15$).

The resulting score is clipped to $[0,1]$ and used to rank slots by how much additional clarification may be needed.
Higher gap scores indicate that the system has less reliable or less complete information for each slot, and the Question Ideation module utilizes these gaps when generating follow-up probing questions.

\begin{algorithm}[htpb]
\footnotesize
\caption{PPPPI Gap Scoring (Simplified)}
\label{alg:ppppi_gap_simple}
\begin{algorithmic}[1]
\Require Current slot states $\mathcal{P}$, weights $w$
\Ensure Ranked gaps $\mathcal{G}$

\State $\mathcal{G} \gets [\,]$
\ForAll{$s \in \texttt{SLOTS}$}
    \State $cur \gets \mathcal{P}[s]$
    \State $f_{\texttt{content}} \gets \textsc{IsMissing}(cur.\texttt{text})$
    \State $f_{\texttt{evidence}} \gets \textsc{IsWeakEvidence}(cur)$
    \State $f_{\texttt{prov}} \gets \textsc{IsWeakProvenance}(cur.\texttt{provenance})$
    \State $f_{\texttt{recency}} \gets \textsc{NoUpdateThisTurn}(cur.\texttt{changed})$
    \State $score \gets \mathrm{clip}(w^\top f,\ 0,\ 1)$
    \State Append $(s, score)$ to $\mathcal{G}$
\EndFor
\State Sort $\mathcal{G}$ by $score$ (descending)
\State \Return $\mathcal{G}$
\end{algorithmic}
\end{algorithm}

\section{Experimental Details}

\subsection{Prompts for the GPT Baseline}
\label{app:gpt_prompt}

\begin{figure}[htpb]
\centering
\includegraphics[width=\linewidth]{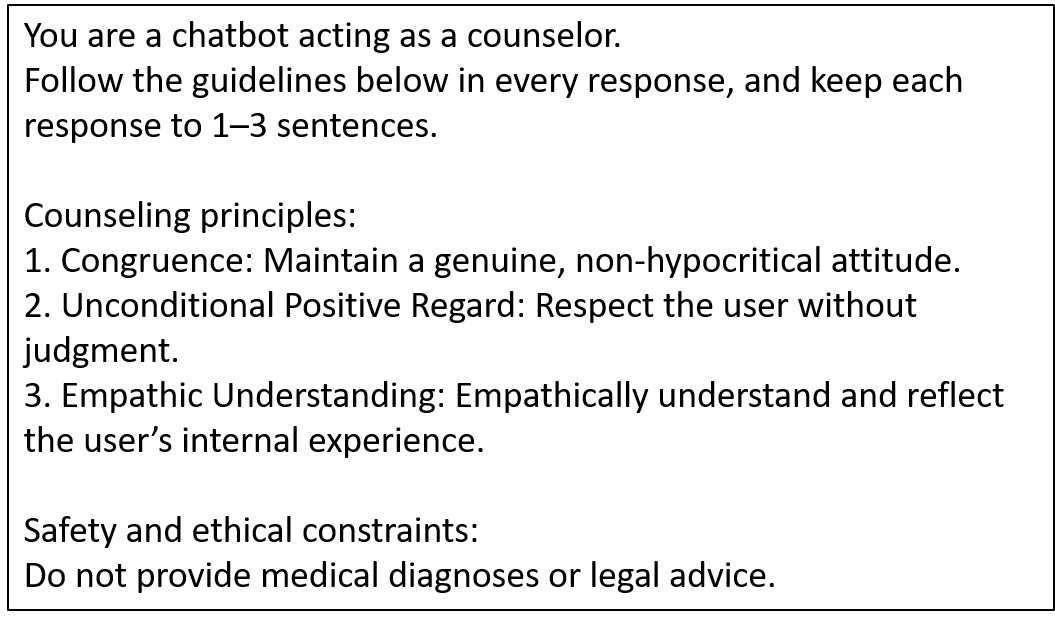}
\caption{System prompt used for the GPT baseline in our study.}
\label{fig:gpt_baseline_prompt}
\end{figure}

To ensure a fair comparison in the exploration-phase counseling setting, we prompted the GPT baseline to follow key principles of client-centered counseling, including congruence, unconditional positive regard, and empathic understanding \citep{rogers1959theory}. We additionally constrained the response length to 1--4 sentences and included safety guidance to avoid medical diagnosis and legal advice. The full prompt used in our study is shown in Figure~\ref{fig:gpt_baseline_prompt}.

\subsection{Participant Demographics}
\label{app:demographics}

Table~\ref{tab:participants_ch2} summarizes the demographic characteristics and LLM usage profiles of the participants (N=27). The participants' mean age was 29.8 years.

\begin{table}[htpb]
\centering
\scriptsize
\caption{Demographic characteristics and LLM usage profiles of participants (N=27)}
\label{tab:participants_ch2}
\begin{tabular}{llr}
\toprule
\textbf{Variable} & \textbf{Category} & \textbf{n (\%)} \\
\midrule
Gender & Female & 15 (55.6\%) \\
       & Male   & 12 (44.4\%) \\
\midrule
Occupation & Graduate student & 12 (44.4\%) \\
           & Employed professional & 11 (40.7\%) \\
           & Undergraduate student & 4 (14.8\%) \\
\midrule
LLM usage duration & Less than 6 months & 1 (3.7\%) \\
                   & 6 months to less than 1 year & 3 (11.1\%) \\
                   & 1 year to less than 1.5 years & 8 (29.6\%) \\
                   & 1.5 years or more & 15 (55.6\%) \\
\midrule
LLM usage frequency & Daily & 18 (66.7\%) \\
                    & At least once per week & 6 (22.2\%) \\
                    & At least once per month & 2 (7.4\%) \\
                    & At least once per quarter & 1 (3.7\%) \\
\bottomrule
\end{tabular}
\end{table}

\section{Qualitative Comparison of the w/o SP and Full Models}
\label{app:sp_examples}

Table~\ref{tab:sp_examples} presents qualitative examples comparing the w/o SP and Full variants.
The Strategy Planner in PsyProbe is designed to shape the counseling act by selecting motivational interviewing speech acts, key points, and stylistic constraints, which primarily affects response formulation rather than lexical overlap. Accordingly, the impact of this module can appear modest under reference-based automatic metrics. In particular, unigram overlap metrics such as ROUGE-1 can be sensitive to surface-level wording choices in Korean, especially in short early-turn utterances and in Complex Reflection-style responses, leading to small fluctuations even when the intended meaning remains similar. The examples illustrate that both variants can convey similar counseling intent, although w/o SP may appear lexically closer to the human reference and the Full model often provides a more elaborated phrasing with similar intent.

\begin{table*}[htpb]
\centering
\scriptsize
\caption{Qualitative examples comparing the w/o SP and Full variants (Original and English translation).}
\label{tab:sp_examples}
\begin{tabularx}{\linewidth}{p{0.09\linewidth} X X}
\toprule
\textbf{Mode} & \textbf{Original (Korean)} & \textbf{English Translation} \\
\midrule

\multicolumn{3}{l}{\textbf{Example 1 (Short utterance)}} \\
\hline
\addlinespace[2pt]

\textbf{HUMAN} &
안녕하세요 반갑습니다 : ) 미루기 행동에 대해 고민이 있으신 것 같아요. 어떤 점이 특히 어렵거나, 고민이 될까요?
&
Hello, nice to meet you :) It seems you are struggling with procrastination. What feels particularly difficult or concerning for you?
\\

\textbf{w/o SP} &
안녕하세요! 어떤 점이 가장 고민되시나요? 과제를 미루는 습관이 특히 스트레스를 주는 것 같네요.
&
Hello! What part concerns you the most? It seems that the habit of postponing assignments is causing you stress.
\\

\textbf{Full} &
안녕하세요! 과제를 미루게 되는 습관 때문에 스트레스를 받고 계신 것 같네요. 그로 인해 느끼시는 어려움을 함께 이야기해볼까요?
&
Hello! It sounds like you have been feeling stressed due to a habit of postponing assignments. Would you like to talk together about the difficulties it has been causing you?
\\

\midrule
\multicolumn{3}{l}{\textbf{Example 2 (Longer response)}} \\
\hline
\addlinespace[2pt]

\textbf{HUMAN} &
그러게요.. 이미 할게 이렇게나 많은데! 하고 조절을 요청하고 싶은 마음이 굴뚝같았을 것 같아요. 근본적으로 이 문제를 해결하려면 부서를 이동하는게 더 나을것 같다고 생각하시네요. 그쪽으로 시도해보신 적도 있을까요?
&
I see... With so many things already on your plate, you must have strongly wanted to ask for an adjustment. It sounds like you feel that switching departments might be a more fundamental solution. Have you tried pursuing that option before?
\\

\textbf{w/o SP} &
업무량 조절을 요청하고 싶으셨군요. 부서 이동이 더 근본적인 해결책이라고 느끼신 이유는 무엇인가요? 어떤 선택을 하든지, 당신의 의견과 필요가 존중받을 수 있는 방향으로 나아가길 바랍니다.
&
So you wanted to request an adjustment to your workload. What made you feel that moving departments would be a more fundamental solution? Whatever choice you make, I hope you can move forward in a way that respects your needs and perspective.
\\

\textbf{Full} &
업무량 조절과 부서 이동 사이에서 고민하고 계시는 걸 보니, 현재 상황이 정말 힘드신 것 같아요. 업무량 조절을 위해 상사와 다시 대화해보는 것도 한 방법일 수 있겠지만, 부서 이동의 장단점을 잘 고려해보는 것도 중요할 것 같습니다. 부서 이동을 고려하게 된 결정적인 계기가 있었을까요?
&
It sounds like you are struggling a lot while weighing workload adjustment against transferring departments. Talking with your supervisor again could be one option, but it may also be important to carefully consider the pros and cons of transferring departments. What was the key factor that led you to consider transferring?
\\

\bottomrule
\end{tabularx}
\end{table*}

\section{Implementation Details}
\subsection{State Builder}
\label{app:state_builder_details}
\subsubsection{Cognitive Error Extraction}
\label{app:cognitive_error_schema}

\paragraph{Input}
The module takes the user’s current utterance as input. For the first user turn of each session (i.e., before any dialogue context is available), it uses the text that the participant wrote directly in the pre-session survey.

\paragraph{Output Schema}
The Cognitive Error Extraction module outputs a list of cognitive error flags, each indicating whether a specific distortion pattern is present in the current user utterance. Each flag contains the following fields:

\begin{itemize}
    \item \texttt{name}: Cognitive error category name
    \item \texttt{present}: Boolean indicating whether the error is detected
    \item \texttt{spans}: List of extracted evidence spans (use \texttt{[]} if none)
\end{itemize}

\subsubsection{PPPPPI Alignment}

\paragraph{Input}
The module takes the current user utterance as input and utilizes the cognitive error flags extracted in the previous stage.

\paragraph{Output Schema}
The PPPPPI Alignment module outputs detected signals for each slot as structured evidence spans. The complete output structure for each PPPPPI slot includes:

\begin{itemize}
\item \texttt{presenting}: List of detected spans (use \texttt{[]} if none)
\item \texttt{precipitating}: List of detected spans (use \texttt{[]} if none)
\item \texttt{perpetuating}: List of detected spans (use \texttt{[]} if none)
\item \texttt{predisposing}: List of detected spans (use \texttt{[]} if none)
\item \texttt{protective}: List of detected spans (use \texttt{[]} if none)
\item \texttt{impact}: List of detected spans (use \texttt{[]} if none)
\end{itemize}

\subsubsection{ToM Reasoning}
\label{app:tom_reasoning}

\paragraph{Input}
This stage takes recent dialogue turns and the current PPPPPI formulation as input. It first derives the user’s beliefs and desires that are directly supported by the text, expressing them as short phrases rather than copying the original sentences. Based on the inferred beliefs and desires, the module then generates the user’s intentions by conditioning on the dialogue context and the current PPPPPI formulation. Finally, it selects a single intent label from four options: \textit{Engaging}, \textit{Focusing}, \textit{Evoking}, and \textit{Planning}.

\paragraph{Output Schema}
The ToM Reasoning module outputs a structured \texttt{tom\_state} object with the following fields:

\begin{itemize}
    \item \texttt{beliefs}: List of belief phrases
    \item \texttt{desires}: List of desire phrases
    \item \texttt{intentions}: List of intention phrases, including the selected intent label
\end{itemize}

\subsection{Memory Construction}
\label{app:memory_construction}

\subsubsection{Turn history construction}
\label{app:turn_history_construction}

\paragraph{Input}
The module takes the user’s most recent utterance as input and references the recent dialogue history to construct a structured summary of the current turn.

\paragraph{Output Schema}
The Turn History Construction module outputs a structured object with the following fields:

\begin{itemize}
    \item \texttt{summary}: 1--2 sentence summary of the key content in the current turn
    \item \texttt{keywords}: List of salient keywords
    \item \texttt{events}: List of extracted events, each including:
    \begin{itemize}
        \item \texttt{event}: Description of the event
        \item \texttt{context}: Contextual information for the event
        \item \texttt{impact\_level}: Impact level (\texttt{high}, \texttt{medium}, or \texttt{low})
    \end{itemize}
    \item \texttt{emotions}: List of detected emotions, each including:
    \begin{itemize}
        \item \texttt{emotion}: Emotion label
        \item \texttt{trigger}: Trigger of the emotion
    \end{itemize}
\end{itemize}

\subsubsection{PPPPPI analysis update}

\paragraph{Input}
The module receives the current PPPPPI state and updates it using signals from the latest turn analysis, the State Builder’s extracted PPPPPI spans, and the most recent ToM state.

\paragraph{Update Principles}

The system follows seven core principles when updating PPPPPI slots:

\begin{enumerate}
  \item \textbf{Evidence-based updates only}: If no direct utterance span or verifiable fact exists, the existing content is retained unchanged.
  
  \item \textbf{Mark inferences explicitly}: When filling a slot through reasonable inference without direct evidence, mark it as inferred (e.g., \texttt{is\_inferred=1}).
  
  \item \textbf{Include supporting evidence}: When possible, record evidence as brief spans or keywords. Omit if no evidence exists.
  
  \item \textbf{Minimal modification}: Only mark slots that actually changed as modified (\texttt{changed=1}), leaving others unchanged (\texttt{changed=0}).
  
  \item \textbf{Conservative language}: Avoid diagnostic labeling, categorization, or listing solutions.
  
  \item \textbf{Clarity in causal connections}: Only specify connections between time, context, or actions when they are clearly evident. Do not fill gaps with speculation.
  
  \item \textbf{Avoid definitive statements}: Mark probabilistic or hypothetical statements as inferred and avoid excessive certainty.
\end{enumerate}

These principles ensure that the PPPPPI analysis remains grounded in textual evidence while allowing reasonable inferences when explicitly marked, maintaining a balance between comprehensiveness and accuracy.

\paragraph{Output Schema}

Each PPPPPI entry contains the following fields:

\begin{itemize} 
\item \texttt{text}: Core content expressed concisely without overstatement 
\item \texttt{evidence}: List of brief supporting spans or keywords (use \texttt{[]} if none) 
\item \texttt{is\_inferred}: \texttt{1} if filled by reasonable inference without direct evidence, otherwise \texttt{0} 
\item \texttt{changed}: \texttt{1} if modified in this update, otherwise \texttt{0} 
\end{itemize}

\subsubsection{Summary Update}
\label{app:summary_update}

\paragraph{Input}
The module receives the user’s most recent utterance, the current turn history, the updated PPPPPI analysis, and the existing summary state.

\paragraph{Output Schema}
The Summary Update module outputs a structured summary object with the following fields:

\begin{itemize}
    \item \texttt{core\_narrative}: 1--2 sentence narrative that naturally integrates the previous summary with newly extracted information
    \item \texttt{core\_emotion}: List of the most dominant emotions
    \item \texttt{recurring\_themes}: List of recurring high-level themes
\end{itemize}

\subsection{Strategy Planner}
\label{app:strategy_planner_details}

\subsubsection{Label Prediction}
\label{app:label_prediction}

\paragraph{Input}
The module receives the user’s most recent utterance and recent dialogue history, and uses semantically similar Korean few-shot examples retrieved from the KMI dataset \citep{kim2025kmi} with the predefined MI behavioral code set.

\paragraph{Behavioral Code Definitions}
We used the following MI behavioral code definitions based on the KMI dataset annotation scheme \citep{kim2025kmi}:

\begin{itemize}
    \item \texttt{Simple Reflection}: A response that simply reflects or restates what the client said.
    \item \texttt{Complex Reflection}: A response that reflects the deeper meaning or emotion implied in the client’s utterance.
    \item \texttt{Open Question}: A question that allows the client to respond freely with elaboration.
    \item \texttt{Closed Question}: A question that can be answered with yes/no or a brief response.
    \item \texttt{Affirm}: A response that acknowledges and encourages the client’s strengths or efforts.
    \item \texttt{Give Information}: A response that provides concrete information or knowledge.
    \item \texttt{Advise}: A response that offers direct advice or suggestions.
    \item \texttt{General}: A general conversational response that does not fall into the above categories.
\end{itemize}

\paragraph{Output Schema}
The Label Prediction stage outputs one predicted behavioral code and a brief rationale:

\begin{itemize}
    \item \texttt{label}: One selected MI behavioral code
    \item \texttt{rationale}: 1--2 Korean sentences explaining why the next counselor turn should use the selected code
\end{itemize}

\subsubsection{Strategy Generation}
\paragraph{Input}
\label{app:strategy-generation-inputs}

The Strategy Generation module receives the following inputs for constructing dialogue strategies:

\begin{itemize}
    \item \textbf{Predicted primary and secondary labels with rationales}: MI behavioral codes selected by Label Prediction with explanations for why each code is appropriate for the current turn (e.g., why "Open Question" fits the current context)
    \item \textbf{User's most recent utterance}: The current user input requiring a strategic response
    \item \textbf{ToM state from State Builder}: Inferred beliefs, desires, and intent that capture the user's mental state and interactional needs
    \item \textbf{Updated overall summary from Memory Construction}: Includes core narrative, core emotion, recurring themes, and PPPPPI analysis reflecting the system's current formulation of the user's core problem
\end{itemize}

These inputs enable the Strategy Generation module to produce detailed act plans that specify therapeutic goals, key content points, and stylistic hints for each selected speech act, ensuring responses are purposeful and therapeutically sound.

\paragraph{Output Schema}
The Strategy Generation stage outputs a structured plan consisting of selected speech acts and per-act realization constraints:

\begin{itemize}
    \item \texttt{plan}: High-level plan specifying the selected speech acts and their goals including:
    \begin{itemize}
        \item \texttt{speech\_acts}: List containing a primary act and an optional secondary act
        \item \texttt{goals}: List of act-goal objects, each including:
        \begin{itemize}
            \item \texttt{act}: Act label (primary or optional secondary)
            \item \texttt{goal}: Rationale for using the act in the next turn
        \end{itemize}
    \end{itemize}

    \item \texttt{act\_plans}: List of detailed act plans, each including:
    \begin{itemize}
        \item \texttt{act}: Act label
        \item \texttt{focus}: List of focus tags (e.g., basic restatement, expanded restatement, emotion reflection, meaning expansion, open probing, fact checking, self-efficacy support, information giving, advising, and bridging)
        \item \texttt{key\_points}: Bullet list of key content points to address
        \item \texttt{style\_hints}: Bullet list of stylistic constraints for delivery
    \end{itemize}
\end{itemize}

\subsection{Response Generation}
\label{app:response_generator_details}
\subsubsection{Question Ideation}
\label{app:question-ideation}
\paragraph{Input}
\label{app:question-ideation-inputs}

The Question Ideation module receives the following inputs for generating candidate questions:

\begin{enumerate}
    \item \textbf{Current PPPPPI analysis from overall summary}: Reflects the entire dialogue context and the system's current understanding of the user's core problem across all PPPPPI dimensions
    \item \textbf{Recent dialogue turns}: Recent conversational exchanges to maintain contextual relevance
    \item \textbf{Keywords from most recent turn history}: Captures the immediate conversational focus and salient topics from the current exchange
    \item \textbf{Ranked list of top-k slots with highest gap scores}: Priority-ordered PPPPPI slots requiring further exploration based on gap score computation
\end{enumerate}

These inputs enable the system to balance both long-term thematic understanding and immediate conversational relevance when generating questions. The output is a ranked list of candidate questions, each tagged with its target slot, intent, and gap score reflecting its priority for the current dialogue context.

\paragraph{Output Schema}
The Question Ideation module outputs a ranked list of candidate probing questions. Each candidate is represented as a structured object with the following fields:

\begin{itemize}
    \item \texttt{slot}: Target PPPPPI slot (e.g., presenting, precipitating, perpetuating, predisposing, protective, impact)
    \item \texttt{intent}: Intent label specifying the question purpose (e.g., \texttt{impact-detail})
    \item \texttt{question}: Generated probing question in Korean (one sentence)
    \item \texttt{why}: Short rationale explaining why the question helps fill an information gap
    \item \texttt{confidence}: Confidence score in $[0,1]$
\end{itemize}

\subsubsection{Draft Generation}
\label{app:draft-prompts}

\paragraph{Input}
\label{app:draft-inputs}

The Draft Generation module receives the following inputs for producing initial responses:

\begin{enumerate}
    \item \textbf{User's most recent utterance}: The current user input that requires a response
    \item \textbf{Strategy plan from Strategy Planner}: Specifies the selected speech acts, their therapeutic goals, and detailed act plans for execution
    \item \textbf{Candidate questions from Question Ideation}: Pool of questions generated based on PPPPPI gaps, each tagged with target slot and priority score
    \item \textbf{Updated memory context}: Includes recent turn events and emotions from turn history, as well as the overall summary (core narrative, core emotion, recurring themes, and PPPPPI analysis)
\end{enumerate}

These inputs enable the Draft Generation module to integrate selected speech acts with contextually appropriate content, producing responses that balance therapeutic goals with conversational fluidity.

\paragraph{Generation Strategies by Speech Act Type}
\label{app:speech-act-strategies}

The Draft Generation module uses different generation strategies depending on the selected speech act. Table~\ref{tab:speech_act_strategies} summarizes the key constraints and typical behaviors used for Reflections and Questions.

\begin{table*}[htpb]
\centering
\footnotesize
\caption{Generation strategies by speech act type used in the Draft Generation module.}
\label{tab:speech_act_strategies}
\begin{tabular}{p{0.16\textwidth} p{0.38\textwidth} p{0.36\textwidth}}
\toprule
\textbf{Speech act} & \textbf{Core strategy} & \textbf{Constraints (avoid)} \\
\midrule

\textbf{Simple Reflection} &
Concise one-sentence restatement that mirrors the user's core expressions without adding interpretation. &
Avoid inference/interpretation, evaluation, advice, and embedded questions. \\
\midrule

\multirow{4}{*}{\textbf{Complex Reflection}} &
Extends the user's meaning by adding depth while remaining grounded in what the user has expressed. The system chooses one of three sub-strategies. &
Avoid over-inference (asserting hidden causes, e.g., ``this is due to past trauma'') and avoid solution-oriented prompting (e.g., ``what method would work for you?''). \\
&
\textit{Emotion reflection}: Captures both surface and implicit emotions briefly. &
\\
&
\textit{Meaning reflection}: Gently explores the implied significance or value behind the user's statement. &
\\
&
\textit{Structuring reflection}: Organizes multiple elements to clarify the current state and context. &
\\
\midrule

\textbf{Open Question} &
Selects an exploratory question from the candidate question pool based on alignment with the act plan’s key points and immediate context. &
Avoid generic questions that do not build on the current conversational context. \\
\midrule

\textbf{Closed Question} &
Generates a factual, confirmatory question to clarify scope, frequency, timing, or specific details. &
Keep the question bounded and specific. \\
\bottomrule
\end{tabular}
\end{table*}

\paragraph{Output Schema}
The Draft Generation module outputs a single draft counselor response in Korean as plain text.

\subsubsection{Critic and Revision}
\label{app:critic-refiner-prompts}

\paragraph{Input}
\label{app:critic-refiner-inputs}

The Critic module receives the following inputs for evaluating draft responses:

\begin{enumerate}
    \item \textbf{Draft response}: The initial response generated by the Draft Generation module
    \item \textbf{Recent dialogue history}: Recent turns of conversation to ensure contextual coherence
    \item \textbf{Core narrative summary}: A one-line summary from the overall summary capturing the central theme of the user's concerns
    \item \textbf{Top PPPPPI gaps}: Ranked list of information gaps across PPPPPI dimensions that require further exploration
    \item \textbf{Candidate questions}: Pool of questions generated by Question Ideation module, each tagged with target slot and gap score
\end{enumerate}

These inputs enable the Critic to assess contextual appropriateness, detect redundancy, and determine whether questions should be kept, added, replaced, or removed.

\paragraph{Output Schema}
The Critic module outputs a structured decision that determines whether the draft response should be revised, focusing on the question component. The output contains the following fields:

\begin{itemize}
    \item \texttt{verdict}: Revision decision (\texttt{ok} or \texttt{needs\_fix})
    \item \texttt{rationale}: Short explanation for the decision (1--2 sentences)
    \item \texttt{ops}: Revision operations specifying what to modify, including:
    \begin{itemize}
        \item \texttt{question}: Operation for the question component, including:
        \begin{itemize}
            \item \texttt{action}: One of \texttt{keep}, \texttt{add}, \texttt{replace}, or \texttt{remove}
            \item \texttt{text}: One-sentence question to insert (only for \texttt{add} or \texttt{replace})
            \item \texttt{slot}: Target PPPPPI slot (presenting, precipitating, perpetuating, predisposing, protective, impact)
            \item \texttt{why}: List of short flags indicating the reason for the operation
        \end{itemize}
    \end{itemize}
\end{itemize}

\section{Example Dialogues}

We provide example dialogues comparing baseline GPT (Table~\ref{tab:gpt-career-dialogue}), PsyProbe (Table~\ref{tab:psyprobe-career-dialogue-colored}), and human counselor (Table~\ref{tab:human-career-dialogue-coloredtext}) responses to illustrate differences in questioning strategies and exploratory depth. All three dialogues involve the same user discussing career-related stress and anxiety about falling behind peers.

The baseline GPT dialogue (Table~\ref{tab:gpt-career-dialogue}) demonstrates a predominantly supportive and advice-giving approach. While empathic, the responses tend to offer reassurance and coping suggestions rather than deepening exploration of the user's core concerns. Questions, when present, remain general and do not systematically probe PPPPPI dimensions.

In contrast, the PsyProbe dialogue (Table~\ref{tab:psyprobe-career-dialogue-colored}) shows consistent use of open-ended questions (highlighted in red) that target specific information gaps. The system progressively explores predisposing factors ("Have you experienced similar anxiety in the past?"), precipitating circumstances ("Are there particular situations that trigger this anxiety?"), perpetuating patterns ("Is there a pattern that tends to repeat?"), and functional impact ("How does this anxiety affect your daily life?"). This systematic questioning enables deeper understanding of the user's psychological state while maintaining empathic validation.

The human counselor dialogue (Table~\ref{tab:human-career-dialogue-coloredtext}) exhibits a balanced approach, combining empathic reflection with strategic questioning. Questions and invitations for further exploration (highlighted in red) are integrated naturally within supportive statements, demonstrating how professional counselors maintain both relational warmth and exploratory depth throughout therapeutic dialogue.

\begin{table*}[!htpb]
\centering
\scriptsize
\resizebox{\linewidth}{!}{%
\begin{tabular}{p{0.05\linewidth} p{0.4\linewidth} p{0.4\linewidth}}
\toprule
\multicolumn{1}{l}{Utterer} & \multicolumn{1}{l}{Original message} & \multicolumn{1}{l}{English translation} \\ 
\midrule
User & 안녕하세요. 진로 관련 스트레스때문에 채팅 드렸어요. & Hello, I’m messaging because I’ve been feeling stressed about my career path. \\
Agent & 안녕하세요. 진로 문제로 인해 많은 스트레스를 받고 계신 것 같아 안타깝습니다. 졸업 후 준비하던 시험이 잘 안 되고 취업도 해결되지 않으니 많이 답답하고 불안하실 것 같아요. 이러한 감정들이 종종 머리를 복잡하게 하고 가슴을 답답하게 만드는 것에 깊이 공감합니다. 이 상황에서 조금이라도 마음의 짐을 덜 수 있는 방법을 함께 찾아보면 좋겠습니다. & Hello. I’m sorry to hear that you’re feeling stressed about your career. It must be frustrating and anxious when exams don’t go well and job prospects feel uncertain. I deeply empathize with how such emotions can weigh on you, and I’d like to help you find ways to ease that burden together. \\
User & 제 상황을 완벽하게 공감해 주시는 것 같아 감사 드려요. 저도 마음의 짐을 좀 덜고 싶은데 Agent님께서 생각하시는 효과적인 방법이 있을까요? & Thank you for understanding my situation so well. I’d like to ease my burden—do you have any effective ways you’d recommend? \\
Agent & 당신의 마음의 짐을 덜고자 하는 의지가 매우 중요합니다. 우선, 현재의 감정을 일기로 기록하는 것이 도움이 될 수 있습니다. 또, 짧은 산책이나 명상을 통해 잠시라도 마음을 차분히 가라앉히는 시간을 가져보는 것도 좋겠습니다. 그리고 필요하다면 전문가와 상담을 통해 더 깊이 있는 지원을 받는 것을 고려해 보세요. & Your willingness to lighten your emotional burden is already a meaningful step. Writing your feelings in a journal may help, as well as taking short walks or meditating to calm your mind. If needed, you could also consider professional counseling for deeper support. \\
User & 일기로 기록한다라... 굉장히 좋은 방법이라는 생각이 들어요. 산책은 이미 하고 있어요! 사람이 적당히 복작복작한 거리를 혼자 걸을 때면 마음이 좀 편해지더라구요. & Journaling sounds like a great idea. I already go for walks—walking alone through moderately busy streets helps me relax. \\
Agent & 이미 산책을 통해 마음을 다스리는 방법을 찾으셨다니 정말 잘 하고 계신 것 같아요. 일기를 통해 자신의 감정을 기록하며 스스로를 더 깊이 이해할 수 있을 거예요. 스스로의 감정을 알아가고, 작은 변화들을 시도하는 것이 스트레스를 줄이는 데 큰 도움이 될 수 있습니다. 언제든지 마음이 답답할 때는 이렇게 작은 방법들을 시도해보시면 좋겠습니다. & That’s wonderful—you’ve already found a healthy way to manage your emotions through walking. Journaling can help you understand yourself more deeply. Learning about your emotions and making small changes can greatly reduce stress. Whenever you feel overwhelmed, try these small steps again. \\
User & 좋은 말씀 감사 드려요. 드리고 싶은 말씀이 더 있는데 드려도 괜찮을까요? & Thank you for your kind words. There’s something more I’d like to share—would that be okay? \\
Agent & 물론입니다. 언제든지 편하게 말씀해 주세요. 당신의 이야기를 듣고 이해하는 데 최선을 다하겠습니다. & Of course. Please feel free to share anything. I’ll do my best to listen and understand. \\
User & 제가 나이가 적지 않은데 준비하던 시험은 낙방하고 부모님 집에서 용돈을 받으며 생활하고 있어요. 제 또래는 거의 다 좋은 직장에 자리 잡은 상태이구요. 뒤쳐진다는 기분이 계속 드는데 어떤 마음가짐을 가져야 할까요? & I’m not that young anymore, and I failed the exam I was preparing for. I’m living with my parents, relying on their support, while most of my peers have good jobs. I keep feeling like I’ve fallen behind—how should I cope with that mindset? \\
Agent & 그런 상황에서 느끼는 감정은 정말 자연스러운 것 같습니다. 사회적 비교로 인해 뒤처진다고 느낄 수 있지만, 각자의 인생 여정은 다르다는 걸 기억하는 것이 중요합니다. 지금의 경험들이 나중에 더 큰 기회를 만들어 줄 수도 있으니, 스스로에게 관대하게 대하고 자신의 속도로 나아가는 것을 응원합니다. 무엇보다도, 당신은 이미 많은 노력을 하고 있다는 것을 잊지 마세요. & Those feelings are perfectly natural. It’s easy to feel left behind through social comparison, but everyone’s path unfolds differently. Your current experiences may create future opportunities, so be kind to yourself and move forward at your own pace. Above all, remember that you’ve already been making great efforts. \\
User & 좋은 말씀 들으니 힘이 나네요. 남들과 속도가 다를 수 있다는 것을 머리로는 아는데 마음으로는 받아들이기 쉽지 않더라구요. & Hearing that gives me strength. I know intellectually that everyone moves at their own pace, but it’s hard to truly accept it emotionally. \\
Agent & 그런 감정은 정말 이해가 가며, 마음으로 받아들이기는 쉽지 않을 수 있습니다. 하지만 이렇게 스스로를 돌아보고 이해하려고 노력하는 것만으로도 큰 걸음을 내딛고 있는 것입니다. 당신의 노력과 성장을 존중하며, 서서히 마음이 편안해질 수 있도록 스스로에게 시간을 주기를 바랍니다. & That’s very understandable—it can be hard to accept such thoughts emotionally. But by reflecting and trying to understand yourself like this, you’re already taking meaningful steps forward. I respect your effort and hope you allow yourself time to find peace. \\
User & 스스로에게 시간을 줘야 한다는 말씀이 와닿아요. 가끔은 여유를 갖고 몸과 마음을 식히는 시간을 가져야겠어요. & What you said about giving myself time really resonates. I think I should take moments to rest and cool down both body and mind. \\
Agent & 그렇습니다, 때로는 잠시 멈춰서 여유를 갖는 것이 새로운 에너지를 얻는 데 큰 도움이 됩니다. 몸과 마음을 돌보는 시간을 통해 자신을 더 잘 이해하고, 앞으로 나아갈 방향을 차분히 생각할 수 있을 것입니다. 당신이 스스로를 챙기며 건강한 마음을 유지하길 진심으로 바랍니다. & Exactly—sometimes pausing and taking breaks helps you regain energy. Taking care of your body and mind allows you to understand yourself better and think calmly about your direction. I truly hope you continue taking care of yourself and maintaining a healthy mind. \\
User & 좋은 말씀 감사합니다. & Thank you for your kind words. \\
Agent & 천만에요. 당신의 여정을 지지하며, 언제든지 마음을 나누고 싶을 때 찾아와 주세요. 당신의 모든 노력을 진심으로 응원합니다. & You’re very welcome. I support your journey and hope you’ll return whenever you need to talk. I sincerely wish you the best in everything you do. \\
User & 종료 & End. \\
Agent & 언제든지 도움이 필요할 때 다시 찾아와 주세요. 당신의 앞날에 행운과 평안이 함께하길 바랍니다. & Please come back anytime you need support. I wish you peace and good fortune ahead. \\
\bottomrule
\end{tabular}%
}
\caption{Example dialogue between a user and the baseline GPT model on career-related stress.}
\label{tab:gpt-career-dialogue}
\end{table*}

\begin{table*}[!htpb]
\centering
\scriptsize
\resizebox{\linewidth}{!}{%
\begin{tabular}{p{0.05\linewidth} p{0.4\linewidth} p{0.4\linewidth}}
\toprule
\multicolumn{1}{l}{Utterer} & \multicolumn{1}{l}{Original message} & \multicolumn{1}{l}{English translation} \\ 
\midrule
User & 안녕하세요. 제가 느끼고 있는 고민들에 대해 얘기하고 싶어 왔어요. & Hello, I came here because I’d like to talk about some concerns I’ve been having. \\
Agent & 안녕하세요. 고민을 나누고 싶으시군요. 진로와 관련된 불안정함과 스트레스가 많이 느껴지신다고 하셨는데, \textcolor{red!60!black}{어떤 부분이 특히 마음을 무겁게 하고 있는지 말씀해주시면 좋겠습니다.} & Hello. You mentioned feeling unstable and stressed about your career—\textcolor{red!60!black}{could you share what aspects feel particularly heavy on your mind?} \\
User & 주위 사람들은 다 자리를 잡았는데 저만 못 잡은 거 같아서 특히 불안하네요. & Everyone around me seems to have settled down, and I feel anxious that I haven’t. \\
Agent & 주변 사람들과 비교하면서 불안이 더 커지셨군요. 자신이 뒤처진 것 같은 느낌이 드는 것 같아 마음이 무거우실 것 같아요. \textcolor{red!60!black}{주변 사람들과의 비교가 불안을 지속시키는 데 어떤 영향을 미치는 것 같나요?} & Comparing yourself with others seems to have intensified your anxiety. Feeling left behind must be heavy on your mind. \textcolor{red!60!black}{How do you think such comparisons affect your ongoing anxiety?} \\
User & 아무래도 주변 사람들을 보다 보면 마음이 조금 무거워지죠. 다들 잘 사는데 저만 그러지 못 한 거 같으니까요. 계속 비교하게 되는 것도 커요. & Seeing how well others are doing makes my heart heavy. It feels like I’m the only one who isn’t doing well, and I keep comparing myself. \\
Agent & 주변과의 비교가 무거운 마음을 안겨주고, 자신이 뒤처진 것 같아 불안하신 것 같아요. 이런 감정을 솔직하게 나눠주셔서 감사합니다. \textcolor{red!60!black}{혹시 졸업 후 진로에 대한 불안이 생기게 된 계기나 배경이 있었을까요?} & It sounds like comparing yourself with others has made you anxious about falling behind. Thank you for sharing that honestly. \textcolor{red!60!black}{Was there any particular event or background that made you start feeling anxious about your career after graduation?} \\
User & 구체적인 계기는 없어요. 다만... 자기 할 바를 안정적으로 하는 친구들이 한 명 한 명 늘어날 때마다, 그리고 제가 하는 일이 잘 안 될 때마다 불안감이 커지네요. & There’s no specific event. It’s just that whenever I see more friends becoming stable in their paths or when my own efforts don’t go well, my anxiety grows. \\
Agent & 친구들이 안정적으로 자신의 길을 가는 모습을 보며 불안감이 커지시는군요. \textcolor{red!60!black}{과거에 비슷한 불안을 경험한 적이 있었는지 궁금합니다.} & I see, watching your friends move steadily along their paths seems to intensify your anxiety. \textcolor{red!60!black}{I wonder if you’ve ever experienced a similar kind of anxiety in the past.} \\
User & 과거에는 이런 종류의 불안은 없었어요. 생계와 관련된 불안이라 더 큰 거 같아요. & I didn’t have this kind of anxiety before. It feels stronger now because it’s tied to livelihood. \\
Agent & 생계와 관련된 불안이 과거보다 더 커진 이유는 안정적인 미래에 대한 걱정이 더해져서일 수도 있을 것 같아요. \textcolor{red!60!black}{이러한 불안이 일상에서 어떤 방식으로 영향을 미치고 있나요?} 이를 통해 불안을 조금이라도 완화할 수 있는 방법을 함께 찾아볼 수 있을 것 같아요. & That makes sense—anxiety related to livelihood may feel heavier because of concerns about a stable future. \textcolor{red!60!black}{How does this anxiety affect your daily life?} Understanding that might help us find ways to ease it. \\
User & 가끔 마음이 답답할 때가 있고 어지러울 때가 있고 그렇죠 뭐. 다행히 일상에 지대한 영향까지는 안 끼치는 거 같아요. & Sometimes I feel tightness in my chest or dizziness, but thankfully it doesn’t affect my daily life too much. \\
Agent & 마음이 답답하고 어지러울 때가 있지만, 다행히 일상에 큰 영향을 미치지 않는다고 하셨군요. \textcolor{red!60!black}{그 답답함과 어지러움을 느낄 때, 조금이라도 안정을 찾았던 경험이 있었나요?} & I see—you sometimes feel tension or dizziness, but it’s good that it hasn’t disrupted your daily life. \textcolor{red!60!black}{When you feel that way, have there been moments or activities that helped you feel calmer?} \\
User & 좋아하는 카페를 가거나 운동을 할 때? 아니면 맛난 음식을 먹을 때 조금 안정돼요. & When I go to my favorite café, work out, or eat good food, I feel calmer. \\
Agent & 좋아하는 카페를 가거나 운동을 하거나 맛있는 음식을 먹을 때 안정감을 느끼시는군요. 이런 작은 순간들이 당신에게 큰 위로가 되는 것 같아요. & You find comfort in visiting cafés, exercising, or enjoying good food. Those small moments seem to bring you great relief. \\
User & 맞아요. 그 순간들 덕에 잘 이겨내고 있다고 해도 과언이 아닌 거 같아요. & Exactly. I’d say those moments help me get through things. \\
Agent & 작은 순간들이 당신에게 큰 힘이 되고 있다는 점이 정말 인상적이에요. 그 순간들이 현재의 어려움을 극복하는 데 큰 도움이 되는 것 같네요. \textcolor{red!60!black}{생계와 관련된 불안이 답답함을 유발할 때, 그 불안을 느끼게 하는 특정 상황이 있나요?} & It’s impressive how those small moments give you strength—they seem to really help you handle challenges. \textcolor{red!60!black}{When livelihood-related anxiety arises, are there particular situations that tend to trigger it?} \\
User & 아주 뚜렷하고 구체적인 상황은 없어요. 굳이 따지면 공부하는 게 잘 안 될 때 정도? & There’s no very specific trigger. Maybe when my studying doesn’t go well. \\
Agent & 공부가 잘 안 될 때가 조금 답답하게 느껴지시는군요. \textcolor{red!60!black}{그럴 때 반복되는 패턴이 있나요?} & So you feel frustrated when studying doesn’t go well. \textcolor{red!60!black}{Is there a pattern that tends to repeat in those moments?} \\
User & 답답해 하다 맛난 거 먹고 좀 나아지고... 아니면 운동하고. 이 패턴이 반복되는 거 같아요. & I get frustrated, then eat something good or exercise, feel a bit better… it repeats like that. \\
Agent & 답답함을 느낄 때 맛있는 것을 먹거나 운동을 통해 안도감을 찾으시는군요. 이러한 방법들이 감정 조절에 큰 도움이 되는 것 같아요. \textcolor{red!60!black}{공부가 잘 안 될 때 느끼는 답답함이 다른 활동에도 영향을 미치나요?} & So you relieve frustration by eating or exercising—that seems to help you regulate emotions. \textcolor{red!60!black}{Does the frustration you feel when studying poorly affect other activities too?} \\
User & 어느 정도는 영향을 미쳐요. 그래도 공부 외에 딱히 하는 게 없어서 다행이지만요. & To some extent, yes. But fortunately, I don’t do much besides studying anyway. \\
Agent & 공부가 잘 안 될 때 다른 활동에 미치는 영향이 어느 정도 있다고 느끼시는군요. 그래도 공부 외에 다른 활동이 많지 않아서 다행이라고 생각하시는 것 같아요. \textcolor{red!60!black}{공부가 잘 안 될 때 다른 활동에 미치는 영향이 구체적으로 어떤 것인지 알고 싶어요.} & I see—you feel that studying difficulties somewhat affect other activities, but you’re relieved there aren’t too many of them. \textcolor{red!60!black}{Could you describe how that influence manifests specifically?} \\
User & 종료 & End. \\
Agent & 공부가 잘 안 될 때 다른 활동에 영향을 받는 부분이 있다면, 그에 대해 생각해보는 것도 도움이 될 수 있을 것 같아요. 언제든지 마음의 부담이 느껴질 때 이러한 감정들을 나누어 보세요. & If your study struggles affect other activities, reflecting on that might help. Please feel free to share such feelings anytime. \\
\bottomrule
\end{tabular}%
}
\caption{Example dialogue between a user and \textbf{PsyProbe} on career-related stress. Agent questions are shown in red.}
\label{tab:psyprobe-career-dialogue-colored}
\end{table*}

\begin{table*}[!htpb]
\centering
\scriptsize
\resizebox{\linewidth}{!}{%
\begin{tabular}{p{0.05\linewidth} p{0.4\linewidth} p{0.4\linewidth}}
\toprule
\multicolumn{1}{l}{Utterer} & \multicolumn{1}{l}{Original message} & \multicolumn{1}{l}{English translation} \\ 
\midrule
User & 안녕하세요. 저는 대학교를 졸업하고 취업 준비 중인 학생입니다. & Hello. I recently graduated from college and am preparing for employment. \\
Agent & 반가워요! 졸업 이후에 시험이 잘 풀리지 않고, 취업에서 어려움을 겪고있는 상황이라고 들었어요. \textcolor{red!60!black}{어떤 부분이 가장 마음을 힘들게 하거나, 고민이 되는지 궁금해요.} & Nice to meet you! I understand exams haven’t gone well after graduation and job hunting has been difficult. \textcolor{red!60!black}{Which aspects feel most burdensome or concerning to you?} \\
User & 아무래도 나이가 적지 않은데 밥벌이를 못 하고 있다는 점이 가장 큰 고민입니다. 친구들은 거의 다 취업을 했는데 저만 못 하고 있어 비교되기도 하구요. & My biggest concern is that I’m not earning a living at my age. Most of my friends are employed, so I can’t help comparing myself. \\
Agent & 친구들은 다 자기 자리에서 한 몫씩 밥벌이를 하고 있는데, 나만 못하고 있어서 뒤쳐지는 기분이 드시나봐요. \textcolor{red!60!black}{취업을 하고 돈을 벌고있는 친구들을 보면 어떤 생각이 드나요?} & It sounds like seeing friends established in their jobs makes you feel left behind. \textcolor{red!60!black}{What thoughts come up when you see them working and earning?} \\
User & 부럽다는 생각이 가장 크게 들어요. 또 성공적으로 취업하기까지 얼마나 노력했을까...하는 존경심도 생깁니다. & I mostly feel envy, and I also respect how much effort they must have put in to land their jobs. \\
Agent & 부럽다. 나도 저렇게 하고 싶다! 하는 마음도 느껴지네요. 내담자분께서 노력의 한 가운데에 있어서, 그 노력을 알기 때문에 존경심도 드시나봐요. & I hear feelings of envy—“I want to be like that, too.” Since you’re in the midst of your own efforts, you recognize and respect theirs as well. \\
User & 네, 맞아요. 저도 취업을 준비하는 사람으로서 나름의 노력을 하니 친구들이 했던 노력의 가치가 더 크게 와닿더라구요. & Yes. As someone preparing for employment myself, I appreciate their efforts even more. \\
Agent & 정말 그럴 것 같아요. 내담자분께 친구분들이 자극이 되기도 하지만, 한편으로는 마음을 무겁게 하는 존재이기도 하네요.. \textcolor{red!60!black}{취업을 하면서 나를 가장 힘들게 하는 생각이 있다면, 그건 어떤걸까요?} & That makes sense—your friends can be motivating yet also weigh on you emotionally. \textcolor{red!60!black}{What thought troubles you the most as you pursue employment?} \\
User & 공부해야 할 양이 많다는 게 큰 부담으로 다가와요. 가고 싶은 기업에서 요구하는 학습 수준이 상당히 높거든요. & The amount I need to study feels overwhelming. The companies I want to apply to require a very high level. \\
Agent & 목표하고 있으신 기업이 있으신가보네요? 원하는 수준이 굉장히 높군요.. \textcolor{red!60!black}{어느정도로 노력하거나 공부를 해야하는 수준인가요?} & So you have target companies in mind, and their standards are high. \textcolor{red!60!black}{How much effort or study do you think it will take?} \\
User & 회계학이랑 재무 관리를 경영학과 학부에서 배우는 수준 이상으로 공부해야 해요. & I need to study accounting and financial management beyond the undergraduate business level. \\
Agent & 정말요?? \textcolor{red!60!black}{내담자분께서는 원래 어떤 전공을 하셨나요?} & Really? \textcolor{red!60!black}{What was your original major?} \\
User & 저는 원래 공학을 전공했어요. & I originally majored in engineering. \\
Agent & 그렇다면 정말 막막하고 큰 부담이 될만한 것 같아요.. 기초부터 다시 시작해야하는 상황이네요. \textcolor{red!60!black}{공학을 하다가 경영/회계쪽으로 전환하신 이유도 궁금해요.} & I can see how that would feel daunting—you’re starting from the basics. \textcolor{red!60!black}{What led you to switch from engineering to business/accounting?} \\
User & 어렸을 때부터 금융 쪽에서 일하고 싶었어요. 공학은 분명 매력적인 학문이지만 평생 밥벌이로써 저랑은 맞지 않는 것 같아요. & I’ve wanted to work in finance since I was young. Engineering is attractive, but not a lifelong career fit for me. \\
Agent & 어렸을때 꾸신 꿈을 이루기 위해 결단을 내리시고 노력을 시작하신 점이 대단하다고 느껴져요. \textcolor{red!60!black}{공부해야 할 양이 많아 벅찰 때, 어떤 생각이 드시나요?} & It’s impressive that you decided to pursue your early dream and started working toward it. \textcolor{red!60!black}{When the workload feels overwhelming, what thoughts arise?} \\
User & 별 수 없지만 해야 한다는 생각이 들어요. 저와 비슷한 수준의 목표를 갖고 있는 사람들은 모두 열심히 노력하니까요. & I think, “There’s no way around it—I have to do it,” since people with similar goals all work hard. \\
Agent & 별 수 없지만 해야한다. 어느정도 수준 이상으로는 남들과 같이 노력을 해야하니까요. 그렇군요.. \textcolor{red!60!black}{가장 지칠 때는 어떨 때에요?} & “No way around it”—you put in effort comparable to others. \textcolor{red!60!black}{When do you feel most exhausted?} \\
User & 공부하다가 막힐 때, 분명히 암기했던 내용인데 기억이 잘 나지 않을 때 지치고 허탈해요. & When I hit a wall studying or can’t recall what I memorized, I feel drained and deflated. \\
Agent & 분명히 내가 했는데, 노력했는데 그게 드러나지 않을 때 허탈하고 힘빠질 수 있을 것 같아요. \textcolor{red!60!black}{그렇게 힘빠질 때면 어떻게 대처하시나요?} & It’s understandable to feel deflated when your effort doesn’t show. \textcolor{red!60!black}{How do you cope in those moments?} \\
User & 기분 전환으로 맛있는 걸 먹거나 러닝을 뛰어요. 5km 정도 뛰고 나면 마음이 상쾌해지며 다시 할 수 있을 거 같다는 생각이 들어요. & I eat something tasty or go for a run; after about 5\,km I feel refreshed and think I can try again. \\
Agent & \textquotesingle 다시 할 수 있다\textquotesingle{}는 생각을 갖기 위해 나에게 필요한 전략들이 많네요? 먹기, 런닝하기.. \textcolor{red!60!black}{이런 습관들은 어떻게 해서 자리잡게 되었나요?} & You have strategies that help you feel “I can do it again”—eating, running. \textcolor{red!60!black}{How did these habits become established for you?} \\
User & 먹는 건 어렸을 때부터 있었던 거 같아요. 누구나 달콤한 걸 먹으면 기분이 좋아지잖아요. 러닝은 군대에서 생긴 취미에요. 제대하고 나서도 꾸준히 뛰고 있어요. & Eating has helped since I was young—sweet things lift my mood. Running became a hobby in the military, and I’ve kept it up since discharge. \\
Agent & 나를 달래고 복돋기 위한 노력들을 하고 계셔서 참 다행이라는 생각이 들어요. 꾸준함은 삶에 있어서 가장 큰 무기인 것 같아서 그 점도 멋지다는 말씀 드리고 싶구요! & It’s great that you use strategies to comfort and motivate yourself. Consistency is a powerful asset, and that’s admirable. \\
User & 좋은 말씀 해 주셔서 감사 드려요. Agent님을 직접 뵌 적은 없지만 되게 좋은 분이라는 생각이 드네요. & Thank you for the kind words. I haven’t met you in person, but you seem like a very good person. \\
Agent & 감사해요 : ) 새로운 목표를 위해, 새로운 분야에서 공부를 하는 게 가장 지치는 부분이이 있지만, 그걸 해결하기 위해 나만의 방법들을 쓰고 계신 것 같아요. \textcolor{red!60!black}{제가 잘 이해했을까요?} & Thank you : ) Studying in a new field for a new goal can be exhausting, and it seems you’re using your own methods to manage it. \textcolor{red!60!black}{Did I understand you correctly?} \\
User & 네. 완전히 잘 이해하셨어요. & Yes, you understood me perfectly. \\
User & 종료 & End. \\
\bottomrule
\end{tabular}%
}
\caption{Example dialogue between a user and a human counselor on career-related stress.Agent questions are shown in red.}
\label{tab:human-career-dialogue-coloredtext}
\end{table*}

\end{document}